\documentclass[11pt]{article}

\usepackage{times}
\usepackage{latexsym}

\usepackage[T1]{fontenc}

\usepackage[utf8]{inputenc}

\usepackage{microtype}

\usepackage{inconsolata}

\usepackage{graphicx}

\usepackage{amsmath}
\usepackage{booktabs}
\usepackage{multirow}
\usepackage{placeins}
\usepackage{subcaption}

\usepackage{booktabs}
\usepackage{multirow}
\usepackage{siunitx}

\usepackage[final]{acl}

\newcommand{\sym}[1]{\rlap{$^{#1}$}}

\title{Estimating LLM Grading Ability and Response Difficulty in Automatic Short Answer Grading via Item Response Theory}

\author{
  \textbf{Longwei Cong\textsuperscript{1}},
  \textbf{Sonja Hahn\textsuperscript{1}},
  \textbf{Sebastian Gombert\textsuperscript{1}},
  \textbf{Leon Camus\textsuperscript{1}}, \\
  \textbf{Hendrik Drachsler\textsuperscript{1,2}},
  \textbf{Ulf Kroehne\textsuperscript{1,3}}
  \\
  \textsuperscript{1}DIPF | Leibniz Institute for Research and Information in Education
  \\
  \textsuperscript{2}Faculty of Computer Science, Goethe University Frankfurt
  \\
  \textsuperscript{3}Chemnitz University of Technology
  \\
  \small{\texttt{\{l.cong,s.hahn,s.gombert,l.camus,h.drachsler,u.kroehne\}@dipf.de}}
}

\begin{document}
\maketitle
\begin{abstract}
Automated short answer grading (ASAG) with large language models (LLMs) is commonly evaluated with aggregate metrics such as macro-F1 and Cohen's kappa. However, these metrics provide limited insight into how grading performance varies across student responses of differing grading difficulty. We introduce an evaluation framework for LLM-based ASAG based on item response theory (IRT), which models grading correctness as a function of latent grader ability and response grading difficulty. This formulation enables response-level analysis of where LLM graders succeed or fail and reveals robustness differences that are not visible from aggregate scores alone. We apply the framework to 17 open-weight LLMs on the SciEntsBank and Beetle benchmarks. The results show that even models with similar overall performance differ substantially in how sharply their grading accuracy declines as response difficulty increases. In addition, confusion patterns show that errors on difficult responses concentrate disproportionately on the \texttt{partially\_correct\_incomplete} label, indicating a tendency toward intermediate-label collapse under ambiguity. To characterize difficult responses, we further analyze semantic and linguistic correlates of estimated difficulty. Across both datasets, higher difficulty is associated with weaker semantic alignment to the reference answer, stronger contradiction signals, and greater semantic isolation in embedding space. Overall, these results show that item response theory offers a useful framework for evaluating LLM-based ASAG beyond aggregate performance measures.
\end{abstract}

\section{Introduction}

Automatic short answer grading (ASAG) is the task of using computational methods to assess short free-text student responses to content questions \cite{burrows2015eras}. Earlier ASAG approaches relied on handcrafted features and classical machine learning models \cite{sultan-etal-2016-fast,mohler-etal-2011-learning}, followed by neural architectures and encoder-based pre-trained language models (PLMs) such as BERT \cite{gombert2023coding,camus2020investigating}. More recently, generative large language models (LLMs) have attracted increasing attention for ASAG because of their strong language understanding and multi-step reasoning capabilities \cite{Cong2026ASAG,frohn2025automated,ferreira2025automatic}.

Despite this progress, evaluation in ASAG remains dominated by aggregate metrics such as accuracy, macro-F1, and agreement coefficients \cite{bonthu2021automated}. While useful for summarizing overall system performance, these measures provide only a coarse view of grading behavior and reveal little about how performance varies across responses of differing grading difficulty \cite{rodriguez-etal-2021-evaluation,choi2026diagnosing}. This limitation is particularly important in ASAG, where student answers are often brief, diverse, and semantically ambiguous \cite{burrows2015eras}. Consequently, two models with similar overall scores may nevertheless differ substantially in how well they handle genuinely difficult responses.

This motivates a more measurement-oriented perspective on LLM-based ASAG. In particular, a useful evaluation framework should help characterize which student responses are difficult to grade and whether LLM graders with similar average performance are equally robust across the difficulty spectrum. Existing ASAG research has paid relatively little attention to these questions at the response level \cite{haller2022survey}.

To address this gap, we draw on item response theory (IRT), a psychometric framework that models observed outcomes as a function of latent ability and item difficulty. Prior work has shown that IRT can support more informative evaluation in NLP when benchmark instances vary substantially in difficulty \citep{lalor-etal-2016-building,lalor-etal-2018-understanding}. We argue that IRT is well suited to ASAG because it separates latent grader ability from response difficulty, enabling a more fine-grained analysis of grading behavior than aggregate metrics alone.

In this work, we apply an IRT-based framework to LLM-based ASAG to study grading performance at the level of individual student responses. Specifically, we make the following contributions:

\begin{itemize}
    \item We introduce an IRT-based perspective for evaluating LLM-based ASAG, estimating latent grader abilities and response grading difficulties across benchmark datasets and multiple LLM graders.
    
    \item We analyze how grading performance changes across response difficulty levels and show that LLM graders differ substantially in their robustness as difficulty increases.
    
    \item We examine semantic and linguistic correlates of response difficulty and find that difficult responses are consistently associated with weaker semantic alignment to the reference, stronger contradiction-related signals, and greater embedding-space isolation.
    
    \item We provide error analyses showing that, on high-difficulty responses, misclassifications are not randomly distributed but increasingly concentrate in the \texttt{partially\_correct\_incomplete} category.
    
\end{itemize}
\section{Background}

\subsection{Evaluation Methods in LLM-based ASAG}

Evaluation in ASAG has traditionally focused on benchmark performance over labeled response sets, typically reported using aggregate metrics such as accuracy, F1, Cohen's kappa, RMSE, and correlation-based measures \cite{haller2022survey,burrows2015eras,sultan-etal-2016-fast}. Recent LLM-based ASAG studies largely follow this paradigm, assessing LLM graders through overall label agreement or classification performance \cite{ferreira2025automatic,frohn2025automated,Cong2026ASAG}.

Yet recent research on LLM evaluation suggests that aggregate benchmark scores alone may provide only a limited view of model capability \cite{zhou2025lost,li-etal-2025-generation}. Prior work shows that benchmark-level averages can obscure substantial variation in item properties such as difficulty and discriminability, while offering limited separability among strong models \cite{lalor-etal-2016-building,lalor-etal-2018-understanding}. Similarly, recent work on LLM-as-a-Judge argues that LLM-based evaluators should themselves be evaluated systematically, with reliability, bias, and robustness as central concerns \cite{gu2025surveyllmasajudge}.

Within ASAG, however, the response-level determinants of grading difficulty remain underexplored, particularly for LLM-based grading. Understanding which properties of student answers make them difficult to grade is important both for interpretability and for deployment, especially in human-in-the-loop settings where difficulty estimates could guide targeted review and improve grading efficiency. Although recent work has examined confidence and uncertainty in ASAG \cite{funayama2022balancing,bexte2024scoring,cong2026confidenceestimationautomaticshort}, these signals are typically model-derived and thus only indirectly informative about the characteristics of the response itself. Analyzing difficulty as a property of the student response instead offers a more interpretable and educationally grounded perspective on the failure modes of LLM-based grading.

\subsection{Item Response Theory}

Item Response Theory (IRT) provides a principled framework by disentangling latent responder ability from item properties \citep{van2016handbook}. In the simplest Rasch formulation, the probability that responder $m$ correctly answers item $n$ is modeled as
\[
P(y_{mn}=1 \mid \theta_m, b_n) = \sigma(\theta_m - b_n),
\]
where $\theta_m$ denotes the latent ability of responder $m$, $b_n$ the difficulty of item $n$, and $\sigma(\cdot)$ the logistic sigmoid. More flexible variants, such as 2PL models, additionally model item discrimination, allowing some items to be more informative than others \cite{hambleton2013item}.

Recent work has begun to apply IRT to LLM evaluation in several ways, including diagnosing the reliability of LLM-as-a-judge systems \citep{choi2026diagnosing} and modeling latent LLM abilities together with query difficulty and discrimination for routing and performance prediction \citep{chen2025learning}. In educational technology, related work has also shown that IRT can be integrated with LLM-based modeling to predict or align item difficulty in assessment settings \citep{scarlatos2025smart}. Collectively, these developments suggest that psychometric modeling offers a robust lens for analyzing LLM behavior through a measurement-grounded perspective.
\section{Method}

\subsection{Datasets and LLMs}

We evaluate our approach on the five-way SciEntsBank and Beetle datasets \cite{dzikovska-etal-2013-semeval}, two widely used benchmarks for ASAG. Released as part of the SemEval student-response analysis task, both datasets contain science-domain questions, reference answers, and student responses annotated by experts. An important property of these benchmarks is their fine-grained five-way label space, which enables analysis beyond binary correctness. Each instance consists of a question, a single reference answer, a student response, and a gold label from the set \texttt{correct}, \texttt{contradictory}, \texttt{partially\_correct\_incomplete}, \texttt{irrelevant}, and \texttt{non\_domain}. An important difference between the two datasets is that Beetle is drawn from a relatively narrow tutorial domain in basic electricity and electronics, whereas SciEntsBank covers assessment responses across a substantially broader set of science topics \cite{dzikovska-etal-2013-semeval}.

As LLM graders, we select 17 open-weight models covering multiple model families including Google, Qwen, Microsoft, Meta-Llama, MistralAI, and OpenChat, with parameter sizes ranging from 0.8B to 14B. The complete list of models is provided in Table~\ref{tab:theta_ranking_merged} in Appendix~\ref{app:plot_table}. The goal is to capture substantial diversity in grading behavior, including variation associated with model size, architectural lineage, and training characteristics.

\subsection{IRT Estimation of Grader Ability and Response Difficulty}

We treat each LLM as a grader and each student response as an item to be evaluated. Given a question, a reference answer, and a student response, each model is prompted to assign exactly one label from the five benchmark categories. Decoding is performed greedily with temperature set to 0 to ensure deterministic and reproducible model outputs.

Let \(i \in \{1,\dots,M\}\) index LLM graders and \(j \in \{1,\dots,J\}\) index student-response instances. For each grader--response pair, we define a binary observation
\[
y_{ij} =
\begin{cases}
1 & \text{if grader } i \text{ is correct on response } j,\\
0 & \text{otherwise.}
\end{cases}
\]

Because multiple student responses are associated with the same question, we include grader-specific testlet effects to capture residual question-specific variation beyond overall grader ability and response difficulty. Specifically, we extend the Rasch model with testlet effects and fit a testlet response theory \cite{wainer2007testlet} variant:
\[
P(y_{ij}=1) = \sigma(\theta_i - b_j + u_{i,t(j)}),
\]
where \(\theta_i\) denotes the latent grading ability of grader \(i\), \(b_j\) denotes the grading difficulty of student-response instance \(j\), and \(t(j)\) denotes the question-level testlet to which response instance \(j\) belongs. The term \(u_{i,t(j)}\) is a grader-specific deviation associated with testlet \(t(j)\), allowing graders to exhibit question-specific performance offsets beyond their overall ability. We use a 1PL testlet formulation rather than a 2PL variant because a 2PL model would introduce an additional discrimination parameter for each response instance. Given the large number of response instances, we therefore adopt the 1PL formulation and account for question-level dependence through testlet deviations.

Parameters are estimated by minimizing the regularized negative log-likelihood:

\begin{equation}
\begin{aligned}
\mathcal{L} ={}&
-\sum_{i,j} \Big[
y_{ij}\log p_{ij}
+ (1-y_{ij})\log(1-p_{ij})
\Big] \\
&+ \frac{\lambda_\theta}{2}\|\theta\|_2^2
+ \frac{\lambda_b}{2}\|b\|_2^2
+ \frac{\lambda_u}{2}\|u\|_2^2 ,
\end{aligned}
\end{equation}
Following our implementation, we set \(\lambda_\theta=\lambda_b=1.0\) and \(\lambda_u=5.0\). To ensure identifiability, \(\theta\) and \(b\) are mean-centered within the fitted dataset, and the testlet effects are centered across graders within each testlet. Optimization is performed using L-BFGS \cite{liu1989limited}.

\subsubsection{Parameter Recovery}

To assess whether the latent parameters are recoverable in practice, we conduct a parameter recovery study using the fitted IRT model as the data-generating process, following standard practice in IRT evaluation \cite{harwell1997analyzing}.
Specifically, we treat the fitted model as a generative data source and conduct a simulation-based parameter recovery analysis. For each replication, we use the estimated grader abilities, response difficulties, and grader-specific testlet effects to compute the probability of correct grading for every grader--response pair under the fitted 1PL testlet model. We then sample a new binary correctness matrix from these probabilities, re-fit the model to the simulated data using the same estimation procedure as for the original data, and compare the recovered parameters with the original fitted parameters after alignment to a common scale.

Recovery is quantified using Pearson correlation, RMSE, and MAE, after aligning the recovered parameters to the original scale via mean and standard-deviation matching. We report these metrics for grader abilities and response difficulties. This analysis assesses whether the inferred latent quantities are sufficiently recoverable to support downstream interpretation.

\subsubsection{Split-Half Stability}

To assess empirical robustness on real data, we conduct a split-half stability analysis. Whereas parameter recovery evaluates recoverability under synthetic responses generated from the fitted model, split-half stability examines whether the inferred parameters remain consistent when the model is re-estimated on disjoint subsets of the observed data.

We use two complementary procedures. To assess grader-ability stability, we randomly split responses into two halves, fit the model separately on each half using the full set of graders, and compare the resulting ability estimates. Because our main model includes question-level testlet effects, response splits are performed within each question. To assess response-difficulty stability, we randomly split graders into two halves, fit the model separately on each subset using the full set of responses, and compare the resulting difficulty estimates.

We repeat both procedures over 10 random replications and quantify stability using Pearson and Spearman correlations, together with RMSE and MAE after mean--standard deviation alignment to account for scale indeterminacy. High split-half agreement indicates that the inferred latent quantities capture reproducible structure in the observed data.

\subsection{Difficulty-Conditioned Analysis of LLM Grader Behavior}

To analyze how grading behavior changes as response difficulty increases, we use the estimated response difficulty \(b_j\) as the primary stratification variable. We partition responses into quantile-based difficulty bins and compute each grader's empirical accuracy within each bin. This stratified evaluation complements aggregate metrics by revealing not only whether performance declines with increasing difficulty, but also whether graders differ in their sensitivity to difficult responses.

Using these bins, we perform two analyses. First, we examine the relationship between difficulty and grading accuracy to assess how performance changes across the difficulty spectrum. Second, we inspect bin-specific confusion matrices to identify how error patterns shift with increasing difficulty.

\subsection{Semantic and Linguistic Correlates of Difficulty}

To examine which response properties are associated with grading difficulty, we extract a focused set of semantic and linguistic features for each student response, motivated by prior work in ASAG \cite{mohler2009text,burrows2015eras,dzikovska-etal-2013-semeval}. These features include response length, lexical diversity, unigram and bigram overlap with the reference answer, semantic similarity to the reference answer, NLI-based features computed on the reference--response pair, embedding-space neighborhood features, and the number of missing reference segments. Sentence-embedding features are computed using \texttt{all-MiniLM-L6-v2} \cite{reimers2019sentence}, and NLI-based features are obtained using a ModernBERT-based NLI model \cite{warner-etal-2025-smarter}. A detailed description of the features is provided in Appendix~\ref{app:feature_definitions}.

We treat these features as correlates of estimated response difficulty. For each dataset, we compute both Pearson and Spearman correlations between each feature and the estimated difficulty \(b_j\), and apply the Benjamini--Hochberg procedure to control the false discovery rate under multiple comparisons \cite{benjamini1995controlling}.

\section{Results}

We report three sets of results. First, we assess the stability of the proposed IRT framework and its latent estimates of grader ability and response difficulty. Second, we examine how LLM grading performance degrades as difficulty increases, including changes in confusion patterns on more difficult responses. Third, we analyze how estimated difficulty relates to semantic alignment and response atypicality.

\subsection{Parameter Recovery and Split-Half Stability}

The fitted IRT model converged successfully on both datasets and produced a broad range of latent grader-ability estimates across the 17 evaluated LLMs, indicating substantial heterogeneity in grading ability. The full model-level results are reported in Table~\ref{tab:theta_ranking_merged} in Appendix~\ref{app:plot_table}.

Parameter recovery results in Table~\ref{tab:recovery} further support the stability of these latent estimates under the fitted model. For grader ability, recovery was nearly perfect on both datasets, with mean aligned Pearson correlations of \(0.999\) on SciEntsBank and \(0.998\) on Beetle, together with very low RMSE/MAE values (\(0.029/0.024\) and \(0.036/0.029\), respectively). Response difficulty estimates were also recovered well, with Pearson correlations of \(0.896\) on SciEntsBank and \(0.895\) on Beetle, and comparable error levels (RMSE \(= 0.505\) and \(0.497\); MAE \(= 0.404\) and \(0.398\)). All recovery runs converged successfully on both datasets.

Complementary split-half analyses on the observed data yielded a similar pattern (Table~\ref{tab:split_half_stability}). Grader ability estimates were highly stable across response halves, whereas response difficulty estimates showed moderate-to-strong stability across grader halves, indicating that the inferred difficulty structure is reproducible but noisier than the model ability scale. Taken together, these results suggest that the IRT framework yields identifiable and empirically robust estimates of both latent grading ability and response difficulty, providing a sound measurement-oriented foundation for the subsequent analyses.

\begin{table}[t]
\centering
\small
\setlength{\tabcolsep}{3pt}
\begin{tabular}{lcccc}
\toprule
Parameter & Pearson $r$ & RMSE & MAE & Conv. \\
\midrule
\multicolumn{5}{l}{\textit{SciEntsBank}} \\
grader ability ($\theta$) & 0.999 & 0.029 & 0.024 & 1.000 \\
response difficulty ($b$)    & 0.896 & 0.505 & 0.404 & 1.000 \\
\midrule
\multicolumn{5}{l}{\textit{Beetle}} \\
grader ability ($\theta$) & 0.998 & 0.036 & 0.029 & 1.000 \\
response difficulty ($b$)    & 0.895 & 0.497 & 0.398 & 1.000 \\
\bottomrule
\end{tabular}
\caption{Parameter recovery results over 10 replications on SciEntsBank and Beetle. Values are mean recovery statistics after mean--std alignment to the ground-truth scale. Conv. denotes convergence rate.}
\label{tab:recovery}
\end{table}

\begin{table}[t]
\centering
\small
\setlength{\tabcolsep}{3pt}
\begin{tabular}{lcccc}
\toprule
Parameter & Pearson $r$ & RMSE & MAE & Conv. \\
\midrule
\multicolumn{5}{l}{\textit{SciEntsBank}} \\
grader ability ($\theta$) & 0.996 & 0.051 & 0.041 & 1.000 \\
response difficulty ($b$)     & 0.791 & 0.595 & 0.461 & 1.000 \\
\midrule
\multicolumn{5}{l}{\textit{Beetle}} \\
grader ability ($\theta$) & 0.994 & 0.055 & 0.043 & 1.000 \\
response difficulty ($b$)     & 0.786 & 0.595 & 0.464 & 1.000 \\
\bottomrule
\end{tabular}
\caption{Split-half stability results over 10 random replications on SciEntsBank and Beetle. Values are mean agreement statistics after mean--std alignment across halves. Conv. denotes convergence rate.}
\label{tab:split_half_stability}
\end{table}

\subsection{Differential Performance Degradation Across Difficulty Levels}

LLM graders differ substantially in how their grading performance degrades as response difficulty increases. We grouped responses into five quantile-based bins according to their estimated difficulty \(b\), from the easiest (\(B1\)) to the hardest (\(B5\)). Across both SciEntsBank and Beetle, accuracy declined monotonically as difficulty increased. At the same time, the rate of decline varied markedly across models, indicating substantial differences in robustness to difficult responses.

On SciEntsBank, all examined models showed sharp performance degradation toward the most difficult bins, but the extent of this decline differed. Gemma-3-12B remained the most robust overall, retaining \(16.6\%\) accuracy on \(B5\), whereas Qwen3.5-9B, Phi-3-medium, OpenChat-3.5, and Llama-3.1-8B fell to \(4.1\%\), \(5.8\%\), \(15.9\%\), and \(2.1\%\), respectively. On Beetle, the same overall pattern held, although the relative robustness profile differed. OpenChat-3.5 degraded more gradually than the other models and retained the strongest performance in the upper-difficulty bins, reaching \(67.2\%\) on \(B4\) and \(12.9\%\) on \(B5\). By contrast, Gemma-3-12B and Phi-3-medium dropped much more sharply after \(B3\), while Llama-3.1-8B approached near-zero performance already by \(B4\).

Additional results for the full set of models are provided in Fig. \ref{fig:acc_by_b_appendix_both} in Appendix \ref{app:plot_table}. Notably, the difficulty--performance slopes are broadly consistent across the two datasets. The cross-dataset association is strong, with a Pearson correlation of \(0.838\) and a Spearman correlation of \(0.706\). This suggests that sensitivity to increasing response difficulty is not merely dataset-specific, but instead reflects a comparatively stable characteristic of LLM grading behavior.

\begin{figure}[t]
    \centering
    \begin{subfigure}[t]{\columnwidth}
        \centering
        \includegraphics[width=\linewidth]{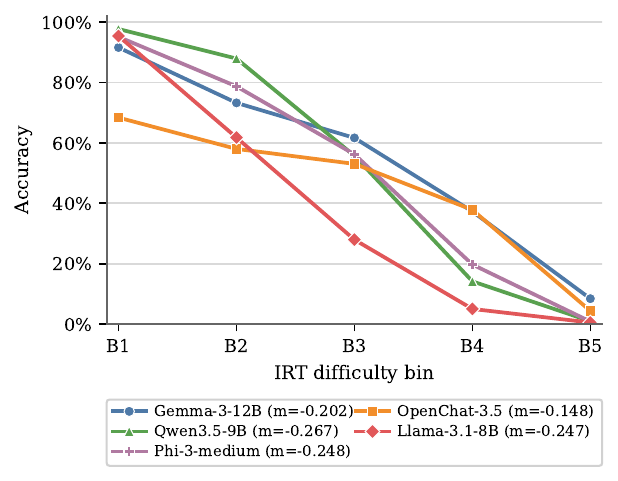}
        \caption{SciEntsBank}
        \label{fig:acc_by_b_scientsbank}
    \end{subfigure}
    \hfill
    \begin{subfigure}[t]{\columnwidth}
        \centering
        \includegraphics[width=\linewidth]{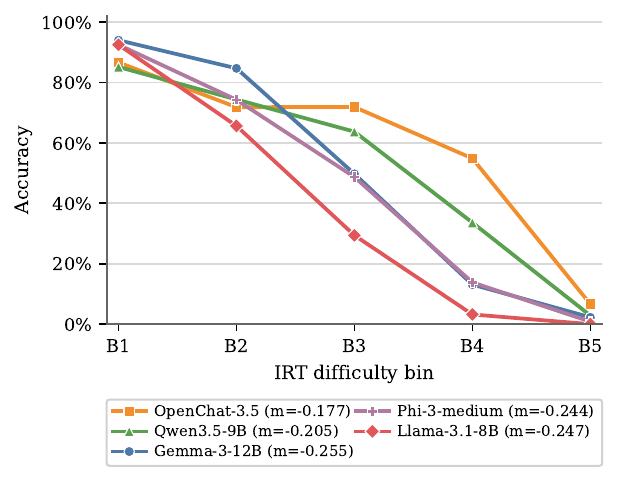}
        \caption{Beetle}
        \label{fig:acc_by_b_beetle}
    \end{subfigure}
    \caption{Model accuracy across ordered response-difficulty bins on (a) SciEntsBank and (b) Beetle. Here, \(m\) denotes the slope obtained by linearly regressing model accuracy on the order of the IRT-based difficulty bins.}
    \label{fig:acc_by_b_bin}
\end{figure}

To further characterize this degradation, we examined confusion patterns across difficulty bins. As shown in Fig.~\ref{fig:scientsbank_confusion_by_b} and Fig.~\ref{fig:beetle_confusion_by_b} in Appendix \ref{app:plot_table}, errors in high-difficulty responses are not randomly distributed across labels, but instead become increasingly structured. Across both datasets, \texttt{correct}, \texttt{contradictory}, \texttt{irrelevant}, and \texttt{non\_domain} responses are increasingly mapped to \texttt{partially\_correct\_incomplete} (PCI) as difficulty rises, indicating a progressive loss of fine-grained multi-class discrimination. At the same time, difficult PCI responses are often misclassified as \texttt{correct}, suggesting that higher difficulty induces both collapse toward an intermediate category and overly optimistic judgments of ambiguous answers.

\begin{figure*}[t]
    \centering
    \includegraphics[width=\textwidth]{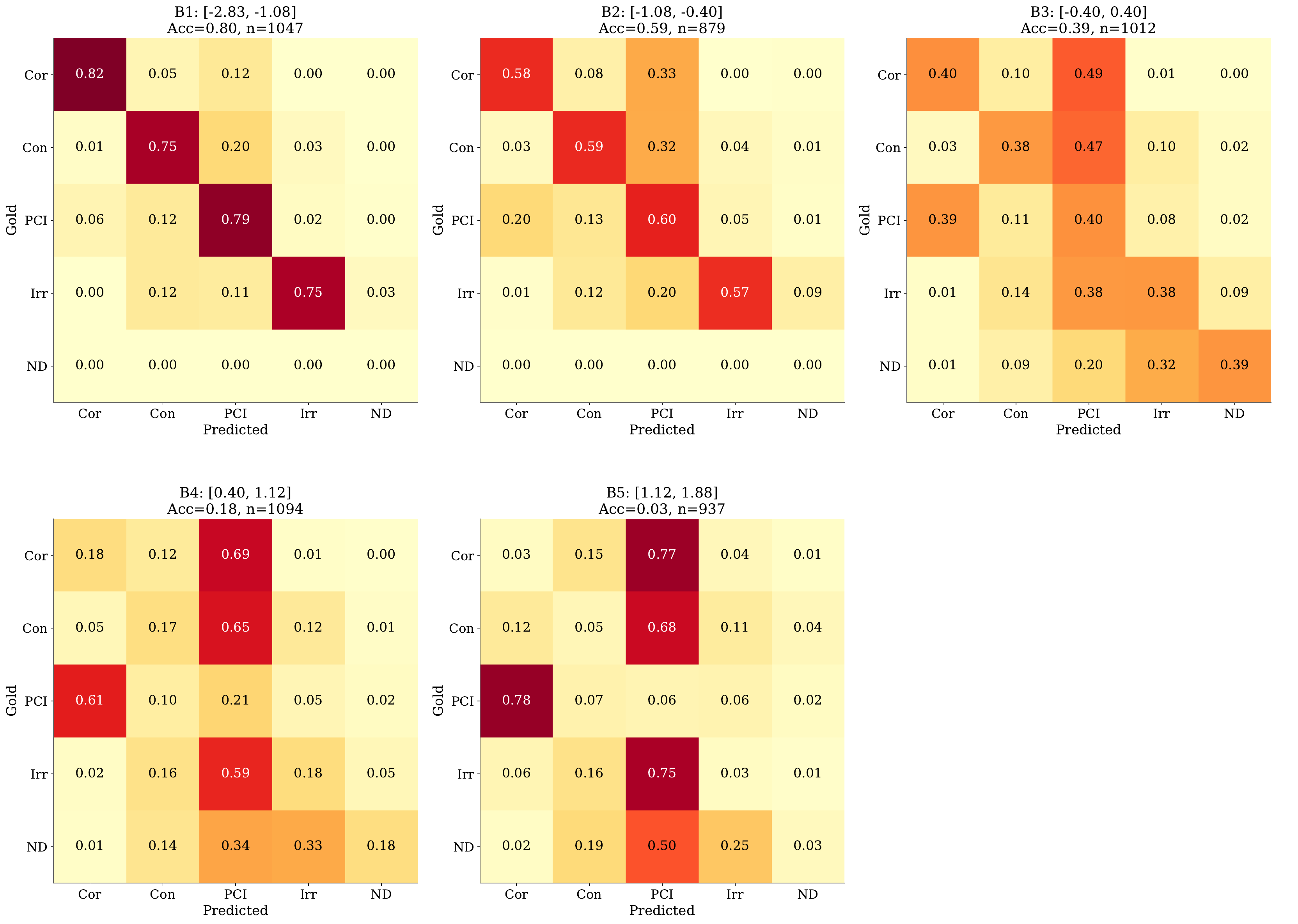}
    \caption{Confusion matrices on SciEntsBank across five bins of IRT-derived response difficulty $b$. $n$ denotes the number of responses in each bin, and Acc denotes overall grading accuracy.}
    \label{fig:scientsbank_confusion_by_b}
\end{figure*}

\subsection{Semantic Misalignment and Response Atypicality as Correlates of Difficulty}

As shown in Table~\ref{tab:feature_corr_cross_dataset}, the correlation analysis reveals a clear and largely consistent pattern across SciEntsBank and Beetle. Overall, the strongest associations with response difficulty come from features that capture semantic alignment with the reference answer and the semantic typicality of a student response relative to other responses. In both datasets, higher difficulty is associated with lower NLI entailment to the reference answer (SciEntsBank: Pearson \(r=-0.300\), Spearman \(\rho=-0.303\); Beetle: Pearson \(r=-0.261\), Spearman \(\rho=-0.227\)) and a smaller entailment--contradiction margin (SciEntsBank: \(r=-0.291\), \(\rho=-0.306\); Beetle: \(r=-0.235\), \(\rho=-0.232\)). Likewise, semantic similarity to the reference answer and lexical overlap with the reference answer, including both unigram and bigram overlap, are negatively correlated with difficulty in both datasets. These results indicate that responses that are semantically closer to the reference answer tend to be easier for LLM graders to evaluate correctly.

By contrast, features reflecting semantic disagreement or atypicality show positive associations with difficulty. In both datasets, the NLI contradiction score is positively correlated with response difficulty (SciEntsBank: \(r=0.211\), \(\rho=0.309\); Beetle: \(r=0.136\), \(\rho=0.216\)). A similar pattern holds for embedding-space neighborhood distance. Both average and minimum \(k\)-nearest-neighbor distance are positively associated with difficulty across datasets. This suggests that responses that are more isolated in semantic space, and therefore less typical relative to the broader response distribution, are more difficult for LLM graders to grade reliably.

At the same time, not all features are equally informative. Some shallow surface-level properties, such as token count and type--token ratio, show weak or inconsistent associations across datasets. The number of missing reference segments is positively related to difficulty in SciEntsBank, but much weaker in Beetle.

\begin{table*}[t]
\centering
\small
\setlength{\tabcolsep}{5pt}
\begin{tabular}{
l
S[table-format=+1.3]
S[table-format=+1.3]
S[table-format=+1.3]
S[table-format=+1.3]
}
\toprule
\multirow{2}{*}{Feature} & \multicolumn{2}{c}{SciEntsBank} & \multicolumn{2}{c}{Beetle} \\
\cmidrule(lr){2-3} \cmidrule(lr){4-5}
& {Pearson $r$} & {Spearman $\rho$} & {Pearson $r$} & {Spearman $\rho$} \\
\midrule
NLI entailment to reference          & -0.300\sym{***} & -0.303\sym{***} & -0.261\sym{***} & -0.227\sym{***} \\
Entailment--contradiction margin     & -0.291\sym{***} & -0.306\sym{***} & -0.235\sym{***} & -0.232\sym{***} \\
Average kNN distance                 & +0.206\sym{***} & +0.250\sym{***} & +0.210\sym{***} & +0.200\sym{***} \\
NLI contradiction to reference       & +0.211\sym{***} & +0.309\sym{***} & +0.136\sym{***} & +0.216\sym{***} \\
Bigram overlap with reference        & -0.247\sym{***} & -0.222\sym{***} & -0.147\sym{***} & -0.110\sym{***} \\
Semantic similarity to reference     & -0.204\sym{***} & -0.220\sym{***} & -0.190\sym{***} & -0.185\sym{***} \\
Minimum kNN distance                 & +0.183\sym{***} & +0.218\sym{***} & +0.187\sym{***} & +0.190\sym{***} \\
Unigram overlap with reference       & -0.203\sym{***} & -0.192\sym{***} & -0.160\sym{***} & -0.127\sym{***} \\
NLI neutrality to reference          & +0.134\sym{***} & +0.162\sym{***} & +0.133\sym{***} & +0.137\sym{***} \\
Number of missing reference segments & +0.133\sym{***} & +0.169\sym{***} & +0.040          & +0.027          \\
Type--token ratio                    & +0.010          & +0.050\sym{*}   & -0.055\sym{***} & -0.050\sym{***} \\
Token count                          & -0.008          & -0.042          & +0.050\sym{***} & +0.064\sym{***} \\
\bottomrule
\end{tabular}
\caption{Correlation between response difficulty and linguistic/semantic features on SciEntsBank and Beetle. Cell values are Pearson $r$ and Spearman $\rho$; superscripts indicate Benjamini--Hochberg adjusted significance ($^{*}q<.05$, $^{**}q<.01$, $^{***}q<.001$). Features are ordered by mean absolute Pearson correlation across the two datasets.}
\label{tab:feature_corr_cross_dataset}
\end{table*}
\section{Discussion}

Our findings support a difficulty-aware perspective on evaluating LLM-based ASAG. Although aggregate measures such as accuracy, macro-F1, and agreement coefficients are useful summaries of overall performance, they do not reveal whether a model remains robust across responses of differing grading difficulty, a limitation also noted in prior IRT-based work on NLP evaluation \cite{lalor-etal-2016-building,lalor-etal-2018-understanding}. The simulation-based recovery and split-half analyses further support the stability of the proposed measurement framework, suggesting that the estimated latent structure is not arbitrary and remains reasonably consistent across repeated analyses. Split-half results also show that grader ability estimates are more stable than response difficulty estimates, likely because the former are estimated from many responses, whereas the latter are supported by far fewer grader judgments. 

This difficulty-aware analysis therefore provides a more informative account of where LLM graders succeed and where they fail. More specifically, the results suggest that LLM graders should be compared not only by their overall performance, but also by how robustly they handle increasingly difficult responses. Models with similar aggregate scores may nevertheless differ substantially in the rate and form of performance degradation across difficulty levels. From this perspective, a grader that performs well on easy cases but deteriorates sharply on difficult responses is qualitatively different from one whose performance declines more gradually, even when their aggregate scores are similar. From a robustness standpoint, the latter may be better suited for practical deployment.

The error analysis suggests that high-difficulty responses do not simply produce more mistakes at random. Instead, errors become increasingly concentrated in the \texttt{partially\_correct\_incomplete} category. This pattern indicates that, under uncertainty, LLM graders often default to an intermediate label rather than making a sharper distinction. One plausible interpretation is that \texttt{partially\_correct\_incomplete} acts as a default intermediate category when evidence is weak, mixed, or only partially supportive. In such cases, the model preserves topical plausibility without committing to full correctness.

The feature analysis helps explain why such responses are difficult. Across datasets, higher difficulty is most consistently associated with weaker semantic alignment to the reference answer, stronger contradiction-related signals, and greater isolation in embedding space. These findings suggest that difficult responses are typically not arbitrary outliers, but responses whose meaning is incomplete, weakly expressed, internally mixed, or only partially compatible with the reference answer. At the same time, the correlations are moderate rather than large, which implies that difficulty is not reducible to any single response property. Instead, grading difficulty appears to emerge from the interaction of multiple semantic factors, including alignment, contradiction, completeness, and local response-space structure.

Taken together, these results have practical implications for LLM-assisted grading. Most importantly, they suggest that not all responses should be treated as equally suitable for fully automatic scoring. If difficulty can be estimated directly or approximated from response features, it could support selective grading pipelines in which straightforward cases are handled automatically while difficult cases are routed to human review. Compared with uncertainty estimates derived solely from the model itself \cite{funayama2022balancing, bexte2024scoring, shorinwa2025survey}, response-based difficulty also offers a more interpretable account of why particular answers are difficult to grade. This may be especially valuable in educational settings, where reliability, interpretability, and fairness are as important as efficiency \cite{madnani-cahill-2018-automated}.

The findings also have implications for rubric design, which has recently emerged as an important direction in ASAG research \cite{frohn2025automated, Cong2026ASAG,gombert2026rubrics}. The systematic tendency to over-predict \texttt{partially\_correct\_incomplete} on difficult responses suggests that current prompting setups may not provide sufficiently clear decision boundaries between neighboring categories. More explicit rubric-based prompting or decomposed decision procedures may help address this problem. For example, grading could be structured into sequential sub-decisions such as topical relevance, presence of correct propositions, contradiction to the reference, and completeness of required content. This kind of scaffolded evaluation may reduce the model's tendency to collapse toward an intermediate class when evidence is ambiguous.
\section{Conclusion}

Overall, the present study shows that IRT offers a useful framework for analyzing LLM graders beyond aggregate performance. Rather than asking only which model achieves the highest average score, this perspective makes it possible to ask which responses are difficult, why they are difficult, and which models remain more robust as difficulty increases. For ASAG, this shift from average performance to measurement-oriented analysis provides a more interpretable basis for both evaluation and system design.

\section{Limitations}

This study has several limitations. First, the number of LLM graders is limited. Although we evaluate 17 models across multiple families and parameter scales, the estimated latent ability scale and robustness patterns should be interpreted with respect to this particular model set. Second, the analysis is restricted to two benchmark ASAG datasets and a specific five-way label schema. Although the main patterns are consistent across SciEntsBank and Beetle, it remains unclear how well they generalize to other domains, rubrics, or grading schemes. Third, our IRT formulation binarizes grading correctness and therefore does not model the five-way label structure directly. Future work could use polytomous IRT models to capture more fine-grained differences among label categories and error types. Finally, the feature analysis is correlational. The observed associations help characterize difficult responses, but they do not establish causal determinants of grading difficulty. Investigating these causal mechanisms remains an important direction for future work.

\section*{Acknowledgments} This research was conducted within the project “Assessment for Learning with AI (ALwAI)” funded by the Leibniz Association under the Leibniz Competition (project no. T163/2024).

\FloatBarrier

\bibliography{custom.bib}

\appendix
\section{Appendix}
\label{sec:appendix}

\subsection{Feature Definitions}
\label{app:feature_definitions}

For each student response, we compute a set of lexical, semantic, and distributional features with respect to the reference answer and the question. Let $a_i$ denote the student answer for item $i$, $r_i$ the corresponding reference answer, and $q_i$ the question. All lexical features are computed from lowercase regex-based tokens extracted with the pattern \texttt{[A-Za-z0-9']+}. Thus, throughout this appendix, ``token'' refers to these surface tokens rather than model-specific subword tokens.

\paragraph{Lexical overlap features.}
Let $U(a_i)$ and $U(r_i)$ denote the sets of unique unigrams in the student answer and reference answer, respectively. The unigram overlap with the reference is defined as
\[
\text{UnigramOverlap}(i) = \frac{|U(a_i)\cap U(r_i)|}{|U(r_i)|}.
\]
Similarly, letting $B(a_i)$ and $B(r_i)$ denote the sets of unique bigrams in the student answer and reference answer, the bigram overlap is
\[
\text{BigramOverlap}(i) = \frac{|B(a_i)\cap B(r_i)|}{|B(r_i)|}.
\]
Both measures therefore quantify reference-normalized lexical coverage rather than Jaccard similarity.

The token count is simply the number of tokens in the student answer,
\[
\text{TokenCount}(i)=|a_i|,
\]
and the type-token ratio is defined as
\[
\text{TTR}(i)=\frac{|\text{unique tokens in }a_i|}{|a_i|}.
\]

\paragraph{Semantic similarity features.}
We encode student answers, reference answers, and questions using Sentence-BERT embeddings. All embeddings are $\ell_2$-normalized, so cosine similarity reduces to a dot product. The semantic similarity to the reference is computed as
\[
\text{SimRef}(i)=\cos(\mathbf{e}(a_i),\mathbf{e}(r_i)).
\]

\paragraph{NLI-based features.}
We apply a pretrained natural language inference model to each reference--answer pair, using the reference answer as premise and the student answer as hypothesis. This yields three probabilities:
\[
P_{\text{ent}}(i),\quad P_{\text{con}}(i),\quad P_{\text{neu}}(i),
\]
corresponding to entailment, contradiction, and neutrality. From these, we derive the following features:
\begin{align*}
\text{NLIEntail}(i) &= P_{\text{ent}}(i),\\
\text{NLIContradict}(i) &= P_{\text{con}}(i),\\
\text{NLINeutral}(i) &= P_{\text{neu}}(i),\\
\text{NLIMargin}(i) &= P_{\text{ent}}(i) - P_{\text{con}}(i).
\end{align*}

\paragraph{Distributional neighborhood features.}
To quantify how typical or atypical a student response is relative to other responses in the same dataset, we compute pairwise cosine similarities among all student-answer embeddings. For each answer, we identify its $k$ nearest neighbors in embedding space (excluding the answer itself), with $k=5$ in our implementation. The average $k$NN distance is defined as
\[
\mathrm{AvgKNNDist}(i)
= 1 - \frac{1}{k}
\sum_{j \in \mathcal{N}_k(i)}
\cos(\mathbf{e}_i, \mathbf{e}_j),
\]
where $\mathbf{e}_i=\mathbf{e}(a_i)$ denotes the embedding of answer $a_i$, and $\mathcal{N}_k(i)$ denotes the set of indices of the $k$ most similar other student answers. The minimum $k$NN distance is defined as
\[
\text{MinKNNDistance}(i)=1-\max_{j\neq i}\cos(\mathbf{e}(a_i),\mathbf{e}(a_j)).
\]
Higher values indicate that a response is more isolated in the embedding space.

\paragraph{Reference-segment coverage.}
To approximate coverage of distinct parts of the reference answer, we split each reference answer into segments using sentence-like punctuation (\texttt{. ; : ! ?}), tokenize each segment, and retain only segments with at least three tokens. For each retained segment $s$, we compute its token overlap with the set of answer tokens:
\[
\text{Coverage}(s,a_i)=\frac{|U(s)\cap U(a_i)|}{|U(s)|}.
\]
A reference segment is considered covered if this value is at least $0.5$. The number of missing reference segments is then
\[
\begin{aligned}
\text{MissingSegments}(i)
&= N_{\text{segments}}(r_i) \\
&\quad - N_{\text{covered}}(r_i) .
\end{aligned}
\]
Higher values indicate that the student answer fails to cover more reference segments.

\onecolumn

\subsection{Additional Plots and Tables}
\label{app:plot_table}

\begin{figure}[!htbp]
    \centering
    \begin{subfigure}{\textwidth}
        \centering
        \includegraphics[width=\linewidth]{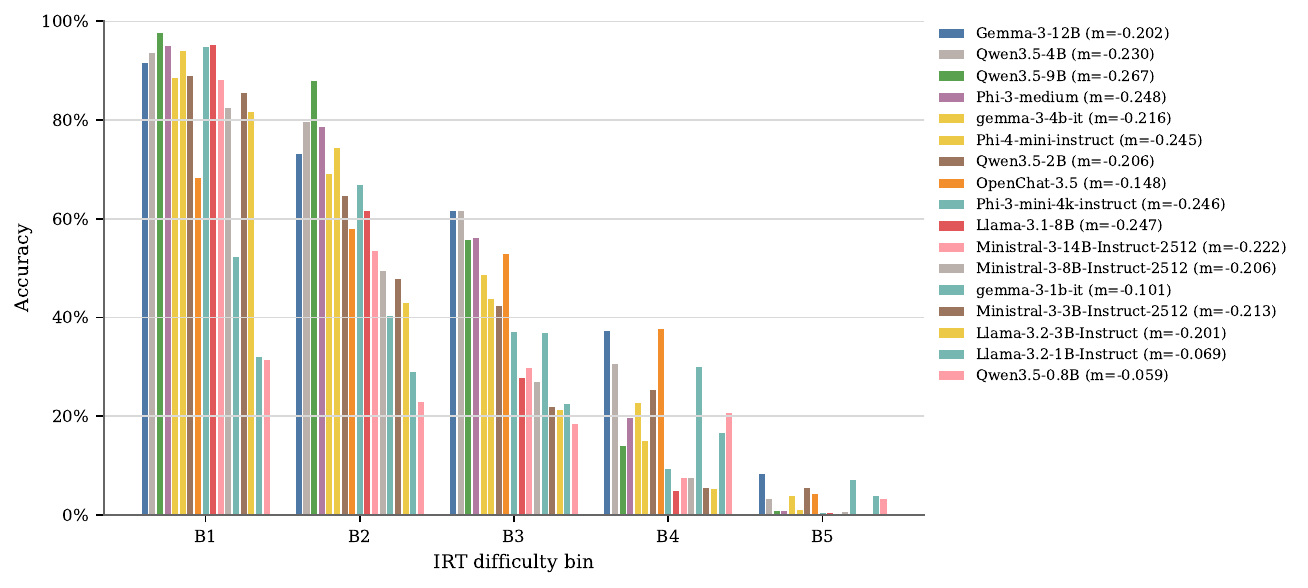}
        \caption{SciEntsBank}
        \label{fig:acc_by_b_appendix_scientsbank}
    \end{subfigure}
    
    \vspace{0.5em}
    
    \begin{subfigure}{\textwidth}
        \centering
        \includegraphics[width=\linewidth]{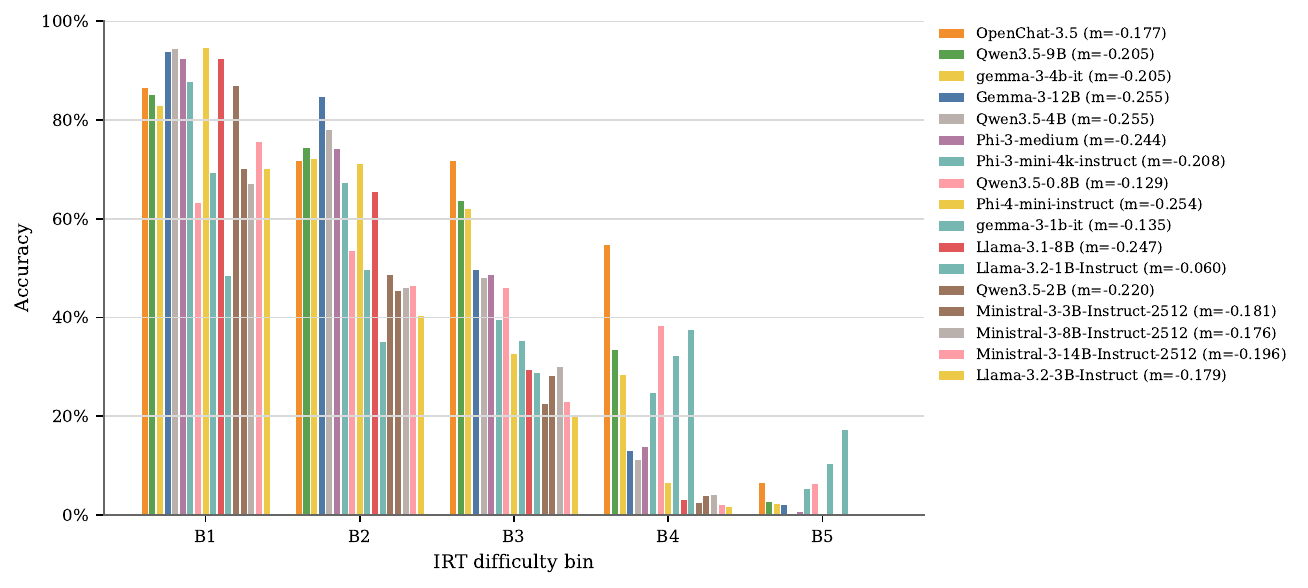}
        \caption{Beetle}
        \label{fig:acc_by_b_appendix_beetle}
    \end{subfigure}
    
    \caption{Model accuracy across ordered IRT-based difficulty bins for all evaluated models on (a) SciEntsBank and (b) Beetle. Here, \(m\) denotes the slope obtained by linearly regressing model accuracy on the order of the IRT-based difficulty bins.}
    \label{fig:acc_by_b_appendix_both}
\end{figure}

\begin{table*}[t]
\centering
\small
\setlength{\tabcolsep}{4pt}
\begin{tabular}{lrrr|rrr}
\toprule
& \multicolumn{3}{c}{SciEntsBank} & \multicolumn{3}{c}{Beetle} \\
\cmidrule(lr){2-4} \cmidrule(lr){5-7}
LLMs & $\theta$ & Acc & Macro-F1 & $\theta$ & Acc & Macro-F1 \\
\midrule
\texttt{google/gemma-3-12b-it}                    & 0.805 & 0.546 & 0.474 & 0.487 & 0.494 & 0.463 \\
\texttt{Qwen/Qwen3.5-4B}                          & 0.758 & 0.538 & 0.434 & 0.404 & 0.470 & 0.314 \\
\texttt{Qwen/Qwen3.5-9B}                          & 0.575 & 0.508 & 0.402 & 0.653 & 0.518 & 0.395 \\
\texttt{microsoft/Phi-3-medium-4k-instruct}      & 0.545 & 0.499 & 0.428 & 0.330 & 0.464 & 0.375 \\
\texttt{google/gemma-3-4b-it}                    & 0.367 & 0.467 & 0.377 & 0.515 & 0.496 & 0.434 \\
\texttt{microsoft/Phi-4-mini-instruct}           & 0.309 & 0.455 & 0.322 & 0.084 & 0.416 & 0.253 \\
\texttt{Qwen/Qwen3.5-2B}                          & 0.285 & 0.455 & 0.350 & -0.463 & 0.322 & 0.172 \\
\texttt{openchat/openchat-3.5-0106}              & 0.264 & 0.446 & 0.356 & 0.963 & 0.577 & 0.494 \\
\texttt{microsoft/Phi-3-mini-4k-instruct}        & 0.117 & 0.416 & 0.335 & 0.197 & 0.450 & 0.414 \\
\texttt{meta-llama/Llama-3.1-8B-Instruct}        & -0.100 & 0.379 & 0.224 & -0.125 & 0.386 & 0.338 \\
\texttt{mistralai/Ministral-3-14B-Instruct-2512} & -0.211 & 0.359 & 0.269 & -0.645 & 0.296 & 0.246 \\
\texttt{google/gemma-3-1b-it}                    & -0.301 & 0.337 & 0.250 & -0.075 & 0.391 & 0.273 \\
\texttt{mistralai/Ministral-3-8B-Instruct-2512}  & -0.349 & 0.335 & 0.242 & -0.649 & 0.297 & 0.282 \\
\texttt{mistralai/Ministral-3-3B-Instruct-2512}  & -0.398 & 0.323 & 0.181 & -0.645 & 0.298 & 0.264 \\
\texttt{meta-llama/Llama-3.2-3B-Instruct}        & -0.491 & 0.304 & 0.148 & -0.798 & 0.266 & 0.161 \\
\texttt{meta-llama/Llama-3.2-1B-Instruct}        & -1.053 & 0.209 & 0.144 & -0.342 & 0.331 & 0.188 \\
\texttt{Qwen/Qwen3.5-0.8B}                        & -1.122 & 0.197 & 0.132 & 0.109 & 0.412 & 0.193 \\
\bottomrule
\end{tabular}
\caption{Latent grading ability estimates ($\theta$), accuracy, and macro-F1 for LLM judges on SciEntsBank and Beetle under the IRT model.}
\label{tab:theta_ranking_merged}
\end{table*}

\begin{figure*}[t]
    \centering
    \includegraphics[width=\textwidth]{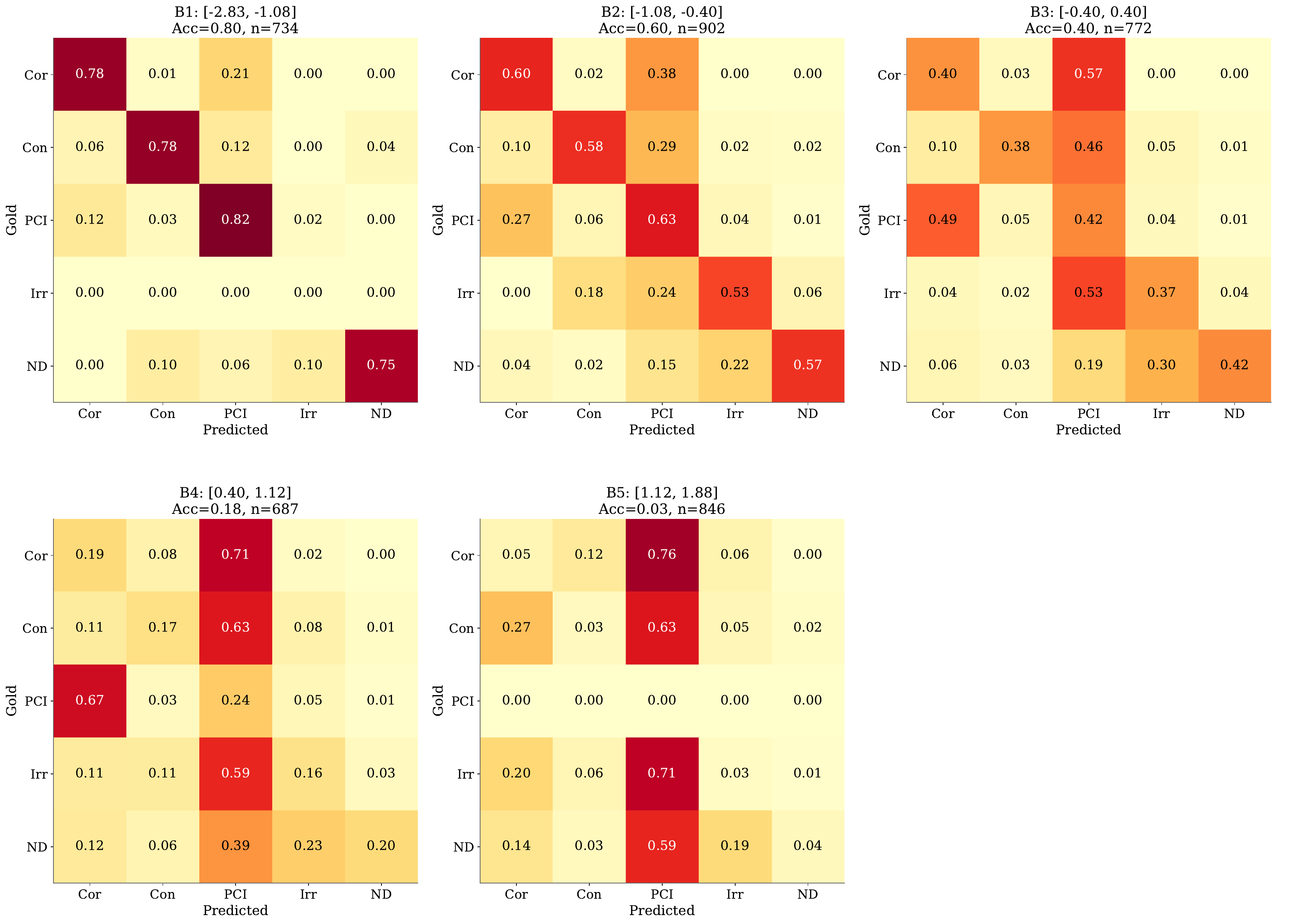}
    \caption{Confusion matrices on Beetle across five bins of IRT-derived response difficulty $b$. $n$ denotes the number of responses in each bin, and Acc denotes overall grading accuracy.}
    \label{fig:beetle_confusion_by_b}
\end{figure*}

\end{document}